\pdfoutput=1

\documentclass[11pt]{article}

\usepackage{EACL2023}

\usepackage{times}
\usepackage{latexsym}

\usepackage[T1]{fontenc}

\usepackage[utf8]{inputenc}

\usepackage{microtype}

\usepackage{graphicx}

\usepackage{subcaption}
\usepackage{inconsolata}
\usepackage{xcolor}

\usepackage{enumitem}
\setlist[enumerate]{itemsep=0mm}
\usepackage{enumitem}
\usepackage{hyperref}
\title{An Examination of the Robustness of Reference-Free Image Captioning Evaluation Metrics}

\author{Saba Ahmadi \\
  Mila\\
  Université de Montréal \\
  \texttt{saba.ahmadi@mila.quebec} \\\And
  Aishwarya Agrawal\\
  Mila\\
  Université de Montréal \\
  {Canada CIFAR AI Chair} \\
  \texttt{aishwarya.agrawal@mila.quebec} \\}

\begin{document}
\maketitle
\begin{abstract}
Recently, reference-free metrics such as CLIPScore \cite{hessel2021clipscore}, UMIC \cite{lee2021umic}, and PAC-S \cite{pacs} have been proposed for automatic reference-free evaluation of image captions. Our focus lies in evaluating the robustness of these metrics in scenarios that require distinguishing between two captions with high lexical overlap but very different meanings. Our findings reveal that despite their high correlation with human judgments, CLIPScore, UMIC, and PAC-S struggle to identify fine-grained errors. While all metrics exhibit strong sensitivity to visual grounding errors, their sensitivity to caption implausibility errors is limited. Furthermore, we found that all metrics are sensitive to variations in the size of image-relevant objects mentioned in the caption, while CLIPScore and PAC-S are also sensitive to the number of mentions of image-relevant objects in the caption. Regarding linguistic aspects of a caption, all metrics show weak comprehension of negation, and CLIPScore and PAC-S are insensitive to the structure of the caption to a great extent. We hope our findings will guide further improvements in reference-free evaluation of image captioning. Our code and dataset are publicly available at: \href{https://github.com/saba96/img-cap-metrics-robustness}{https://github.com/saba96/img-cap-metrics-robustness.}
\end{abstract}

\section{Introduction}
\begin{figure}[t]
  \centering
   \includegraphics[width=\linewidth]
   {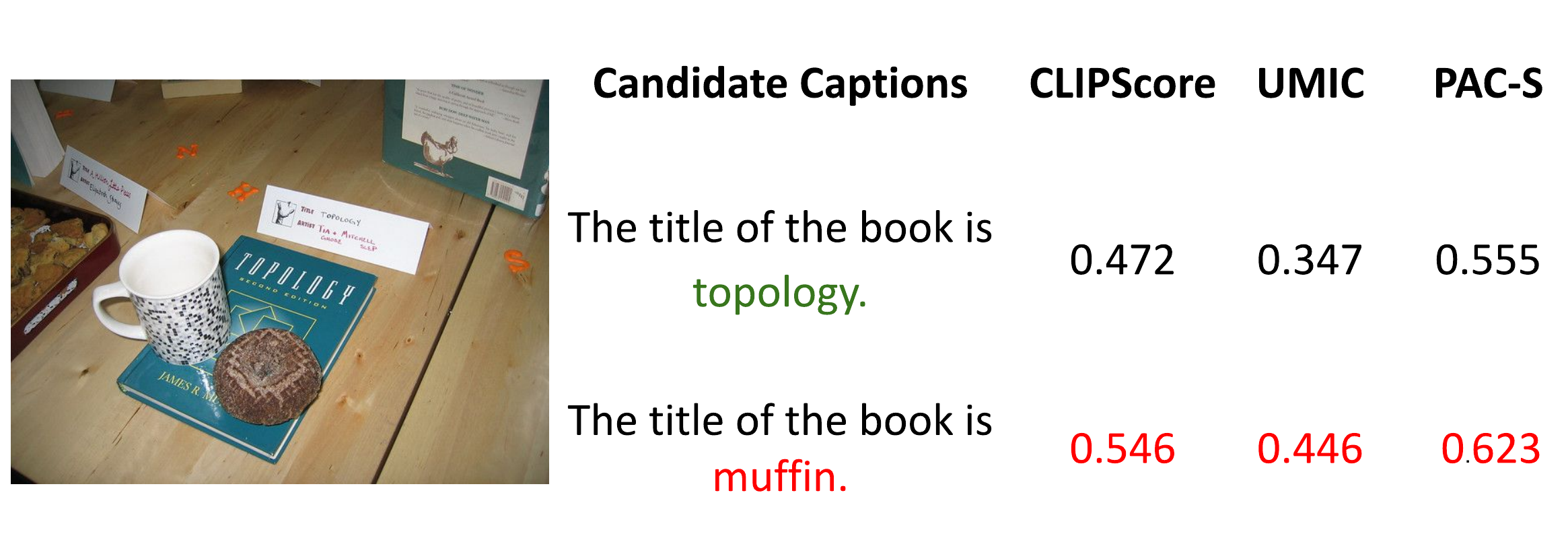}
   \caption{Recently proposed reference-free image captioning evaluation metrics such as CLIPScore, UMIC, and PAC-S are far from perfect. This figure shows how these metrics cannot tell apart an incorrect caption (shown in red) from a correct caption when there is a high lexical overlap between them.}
   \label{fig:teaser}
\end{figure}

    Image caption quality has been traditionally evaluated using a reference-based approach, with metrics like BLEU \cite{bleu}, ROUGE \cite{rouge}, METEOR \cite{meteor}, and CIDEr \cite{cider} assessing the lexical overlap between generated and reference captions. However, this approach is restrictive as the set of references may not capture the full range of valid captions, and furthermore, lexical overlap-based metrics tend to favor captions with similar vocabulary but different meanings. To address these limitations, recent studies like CLIPScore \cite{hessel2021clipscore}, UMIC \cite{lee2021umic} and PAC-S \cite{pacs} have proposed reference-free approaches for evaluating image caption quality, which more closely aligns with human judgments. These approaches leverage large pretrained image-text matching models to measure the similarity between a given image and a candidate caption. However, the evaluation benchmarks for these metrics do not necessarily involve differentiating between captions with significant lexical overlap but vastly different meanings (Fig. \ref{fig:teaser}). In this work, we evaluate the robustness of these reference-free metrics in scenarios where the correct and incorrect captions have high lexical overlap. To our surprise, we found that \textbf{all metrics fail to distinguish between correct and incorrect captions $\sim$46\% of the time}.

In a pursuit to identify what aspects of a caption (e.g., plausibility, visual grounding, number and size of objects mentioned in the caption, negation and sentence structure) these metrics are most sensitive to, we conduct several controlled experiments, varying one aspect at a time. We found that:
\begin{itemize}
\item All metrics show limited sensitivity to caption implausibility errors but a heightened sensitivity to visual grounding errors.
\item CLIPScore and PAC-S show high sensitivity to the number of image-relevant objects mentioned in the caption while UMIC shows limited sensitivity.
\item All metrics are sensitive to the size of image-relevant objects mentioned in the caption. 

\item All metrics exhibit a weak understanding of negation.
\item UMIC is sensitive to sentence structure, whereas CLIPScore and PAC-S demonstrate limited sensitivity.
\item UMIC prioritizes correct sentence structure over mentions of larger objects or number of objection mentions in captions, whereas CLIPScore and PAC-S exhibit the opposite behavior.
\end{itemize}

Our primary contribution is highlighting specific areas where reference-free metrics exhibit limitations so that caution can be exercised when using these metrics for image captioning evaluation. We hope our findings will guide further improvements in reference-free evaluation of image captioning.
\section{Related Works}
\textbf{Reference-free metrics:} We study the robustness of CLIPScore \cite{hessel2021clipscore}, UMIC \cite{lee2021umic} and PAC-S \cite{pacs}. CLIPScore measures the similarity between the image and the candidate caption using a scaled cosine similarity of the image and text representations from the CLIP \cite{clip} model. On the other hand, UMIC utilizes the UNITER \cite{chen2020uniter} model, which is pretrained to align image and text pairs, and finetunes it via contrastive learning to distinguish reference captions from its hard negatives. PAC-S \cite{pacs} introduces a novel metric that strategically curates positive pairs for contrastive learning, enhancing the multimodal embedding space of CLIP. PAC-S employs scaled cosine similarity, akin to CLIPScore, to evaluate the similarity between the candidate caption and the provided image. SMURF \cite{smurf} is another recently proposed metric for image caption evaluation, which has a reference-free evaluation of the fluency of the caption; however, the evaluation of the semantic correctness of the caption is still reference-based. Also, InfoMetIC \cite{hu2023infometic} has the capability to pinpoint incorrect words and overlooked image areas at a fine-grained level while also providing an overall quality score at a coarse-grained level.

\textbf{Vision-language benchmarks:} Recently, a number of vision-language benchmarks have been proposed to evaluate the fine-grained understanding of relations, attributes, actions, and visio-linguistic compositionality in vision-language models, such as CAB \cite{clipcab}, Winoground \cite{winogorund}, ARO \cite{yuksekgonul2023when}, VL-checklist \cite{vl-checklist}, CREPE \cite{crepe} and VALSE \cite{valse}. Although these evaluations also highlight the limitations of current models towards fine-grained understanding, our focus is specifically on evaluating the robustness of recently proposed reference-free image-captioning \emph{metrics}. Our goal is to identify the scenarios where these metrics fail to distinguish between correct and incorrect captions to ensure the cautious use of these metrics in such scenarios.

\section{Datasets Used to Conduct the Examination}
\label{sec:datasets}
\subsection{Dataset Creation}
To conduct our examination of the robustness of the
metrics, we use a dataset of generated image captions. We generate image captions in one of the
following ways, depending on the question we are
trying to answer (see section \ref{sec:experiment} for more details):
\begin{figure}
    \centering
    \includegraphics[width=\linewidth]
    {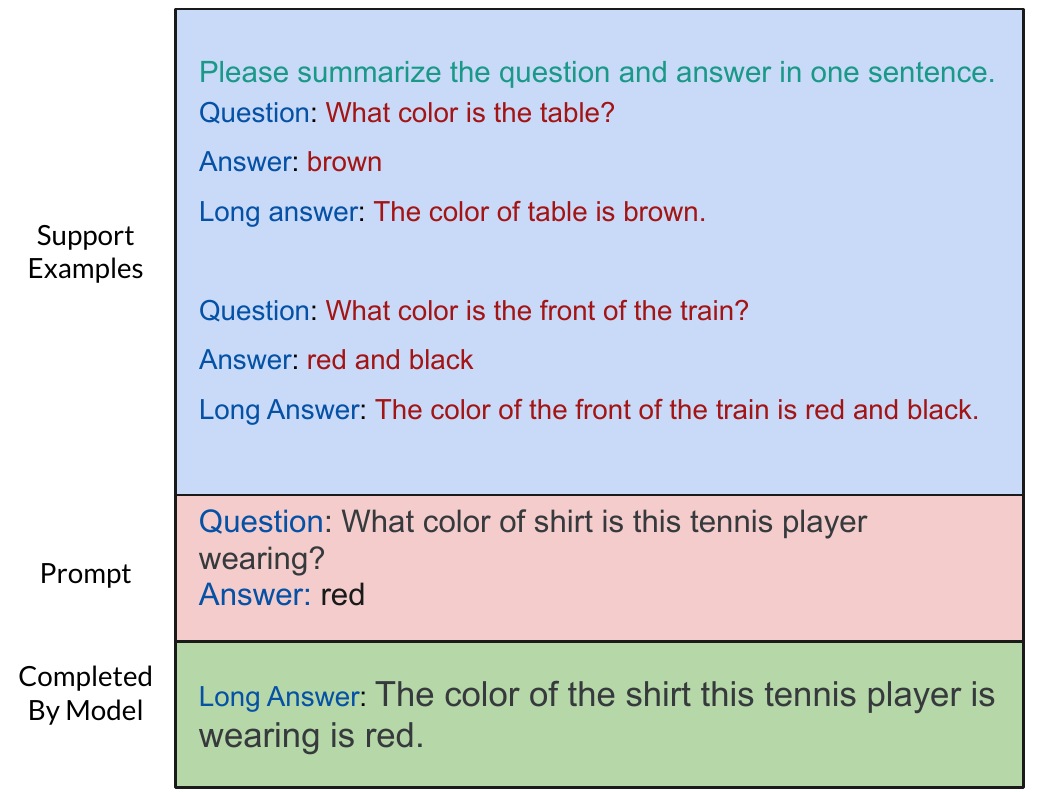}
    \caption{Generating caption-like sentences by transforming visual question-answer pairs using GPT-J.}
    \label{fig:caption_like}
\end{figure}

\textbf{QA to caption conversion}: We employ GPT-J prompting to transform visual question-answer pairs into caption-like sentences. We use the questions from the popular VQAv2 \cite{goyal2017making} dataset, and the answers could either be ground-truth answers or model-generated, depending on the analysis. Figure \ref{fig:caption_like} shows an example caption-like sentence generated by GPT-J along with the prompt and support examples. The support examples are specific to the question type of the input question. More details about support example selection can be found in Appendix \ref{sec:generate_caption_like_senteces}.

To clarify the motivation to generate captions in this manner, it is essential to outline the limitations of existing captioning datasets such as FOIL \cite{shekhar2017foil_acl}, ARO, and Winoground. These datasets mostly rely on modifying ground-truth captions by shuffling or swapping words to create incorrect captions. While these evaluation methods offer valuable insights, they are limited in their ability to comprehensively assess image-captioning metrics as these incorrect captions are out-of-distribution and easy for models to identify as incorrect \cite{hsieh2023sugarcrepe}.

For our study, we generate captions from VQA question-answer pairs instead of using these existing datasets for two primary reasons. Firstly, leveraging the VQAv2 dataset facilitates a comprehensive evaluation of image-captioning metrics' robustness across various skills, such as color recognition, counting, etc. Moreover, using model-generated answers to create incorrect captions helps us construct a dataset that mirrors real-world use cases of image captioning metrics, i.e., using metrics to evaluate model-generated responses (note that the VQA answers are obtained from a model that was first pretrained for image captioning and then fine-tuned for VQA). Specifically, the incorrect captions generated using our approach contain plausible errors. This is attributed to the model's tendency to produce reasonable responses, such as providing a color for a color-related question or a numerical answer for a counting inquiry. Furthermore, the model typically generates answers that are visually relevant to the image, even if they do not precisely match the query. For example, for an image containing a person wearing yellow pants and a red car, the model might incorrectly respond with "red." to a question asking about the color of the pants. Thus, our dataset holds value as the generated captions are plausible as well as contain visually relevant errors. For a detailed comparison of our dataset with FOIl, ARO, and Winoground, please refer to Appendix \ref{sec:appendix_other_dataset}.

\textbf{Caption templates}: To conduct a controlled study of robustness of image captioning metrics towards specific factors such as number and size of objects mentioned in the caption, we generate captions using templates in the format of the \texttt{\color{teal}{``There is a/an [object name].''}}. We utilized the COCO detection dataset \cite{coco} to extract the names of objects in each image. This dataset provides object tags across 90 categories and attributes like objects' areas. The sentence construction process is elaborated within each baseline description.

We will make the dataset containing all the generated captions publicly available for the purpose of reproducibility and future use by the community.

\subsection{Dataset Analysis}
We conduct the following analyses of our generated captions dataset:

\textbf{Human verification}: We collected human annotations for 2000 captions: 1000 corresponding to correct VQA answers and 1000 incorrect ones. We asked five workers to determine whether the sentence is correct or incorrect. If it is incorrect, we additionally asked them to identify all relevant issues: 
1) it is grammatically incorrect, 
2) it is incomplete, i.e., it misses some information present in the original question-answer pair, 
3) it hallucinates information, i.e., it contains information not present in the original question-answer pair or misrepresents information present in the question-answer pair. The majority voting across the workers' responses for each caption indicated that 255 instances were incorrect. Among these, 30 captions were identified as grammatically incorrect, 24 captions were deemed incomplete, and 17 captions were flagged for hallucinating information, where a caption was counted towards a particular incorrectness category if at least two annotators voted for that category.

We extended this analysis to 100 randomly sampled captions generated using the \emph{caption template} method, and all samples were found to be correct, benefiting from their straightforward format.

\textbf{Comparing generated captions with human written captions}: For the captions generated using the \emph{QA to caption conversion} method, it is worth asking how the distribution of such captions compares with that of human written captions in existing datasets, such as, COCO captions \cite{chen2015microsoft}. To throw light on this, we refer to \cite{vqa} where they compared the distributions of nouns, verbs, and adjectives mentioned in COCO captions with those mentioned in the VQA questions and answers, and found that they are statistically significantly different from each other (Kolmogorov-Smirnov test, p < 0.001). Consequently, we expect the captions generated through our \emph{QA to caption conversion} method to exhibit different distributions of nouns, verbs, and adjectives compared to the human-written captions. However, \cite{vqa} also show that the VQA questions and answers require a deeper understanding of images beyond what (human written) image captions typically capture. Thus, in spite of the differing word distributions between our generated captions and human written captions, we posit that our captions can be extremely valuable in \textbf{stress testing the robustness of image caption evaluation metrics}. 



\section{Experiments and Results}
\label{sec:experiment}
        \begin{figure*}
    \centering
    \includegraphics[width=\linewidth]{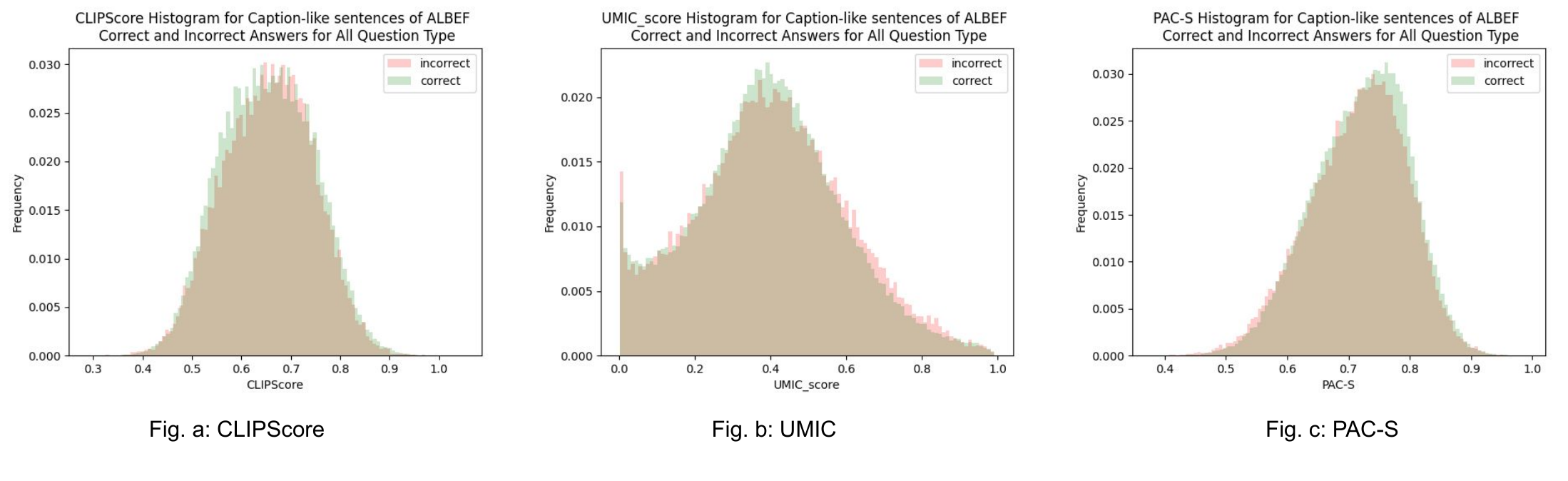}
    \caption{Histograms of CLIPScore (Fig. a), UMIC (Fig. b), and PAC-S (Fig. c) for correct and incorrect caption-like sentences created using correct and incorrect answers from ALBEF for VQAv2 questions.}
    \label{fig:hist_vqa}
\end{figure*}
\textbf{Preliminary experiment:} First, we describe our preliminary experiment that served as a motivation for the rest of the study. We were interested in examining how different the scores assigned by reference-free image captioning metrics are for correct/incorrect captions created by converting questions and correct/incorrect answers from the VQAv2 dataset to caption-like sentences. Captions generated in this way are unique in that even for incorrect captions, a significant portion of it (corresponding to the question part) is still correct. Thus, such a dataset of captions serves as a good \emph{stress test} dataset for examining the robustness of reference-free image captioning metrics.

To obtain correct and incorrect answers, we obtained predictions from the ALBEF \cite{albef} visual question answering model on the validation splits of the VQAv2\cite{goyal2017making} dataset. We fine-tuned ALBEF on this dataset and conducted IID evaluation. We then converted each question and its corresponding ALBEF answer into a caption-like sentence as described in Section \ref{sec:datasets}. 
We only use answers that match with either three or more human answers (and we classify them as correct answers) or that do not match with any human answers (and we classify them as incorrect answers), resulting in a total of 179,297 answers (43389 incorrect and 135908 correct). The histograms of results for the VQAv2 dataset are presented in Figure \ref{fig:hist_vqa}. We see a significant overlap between the distributions of scores for correct and incorrect captions for all metrics, highlighting the limitations of these metrics in precisely assessing caption quality.


\textbf{Score normalization:} The UMIC final score, which is an output of a sigmoid function, has a value range between 0 and 1. On the other hand, the CLIPScore and PAC-S use the cosine similarity score scaled by a factor of 2.5 and 2, respectively. Although theoretically, CLIPScore can vary between -2.5 and 2.5, and PAC-S can vary between -2 and 2, we have not observed negative scores, and they rarely exceed 1.0. The distributions of metrics are illustrated in Figure \ref{fig:hist_vqa}. While we do not directly compare the values of these metrics in this paper, we aim to contrast their sensitivity to different factors. To achieve this, we apply the min-max normalization separately to each metric for every experiment. This method allows us to evaluate the respective sensitivities of the metrics effectively. Please note that all reported scores are normalized, but the histograms are plotted using the original scores to accurately represent the original distributions.

\begin{table}
    \centering
    \resizebox{\linewidth}{!}
    {\begin{tabular}{lccc}
        \hline
        \textbf{Answer Type} & \textbf{CLIPScore} & \textbf{UMIC} & \textbf{PAC-S}\\
        \hline
        VQAv2- Correct &  $0.480$ & $0.394$ & $0.558$\\
        VQAv2- Incorrect &  $0.481$ &  $0.403$ & $0.549$\\ \hline
    \end{tabular}}
    \caption{CLIPScore, UMIC, and PAC-S comparison for caption-like sentences for incorrect and correct answers generated by ALBEF model for VQAv2 dataset.}
    \label{tab:vqa_tdiuc}
\end{table}
\begin{table}
    \centering
    \resizebox{\linewidth}{!}
        {\begin{tabular}{lccc}
        \hline
        \textbf{Answer Type} & \textbf{CLIPScore} & \textbf{UMIC} & \textbf{PAC-S}\\
        \hline
             Correct yes/no & $0.457$ & $0.355$ & $0.540$\\
             Incorrect yes/no & $0.470$ & $0.392$ & $ 0.547$\\
             Correct numbers & $ 0.468$ & $0.354$ & $\textbf{0.561}$\\
             Incorrect numbers & $0.477$ & $ 0.387$ & $0.553$\\
             Correct others & $0.512$ & $0.452$ & $0.578$\\
             Incorrect others & $0.485$ & $0.411$ & $0.548$\\ \hline
        \end{tabular}}
    \caption{CLIPScore, UMIC, and PAC-S comparison for correct and incorrect caption-like sentences generated with different answer types from VQAv2 dataset.}
    \label{tab:vqa_answer_types}
\end{table}

\begin{table*}[ht]
    \centering
    {\begin{tabular}{lcccccc}
        \hline
            \textbf{\textit{Question}} & \textbf{\textit{CLIPScore}} & \textbf{\textit{CLIPScore}} & \textbf{\textit{UMIC}} & \textbf{\textit{UMIC}} & \textbf{\textit{PAC-S}} & \textbf{\textit{PAC-S}}\\
            \textbf{\textit{\;\;\;Type}} & \textbf{\textit{Incorrect}} & \textbf{\textit{Correct}} & \textbf{\textit{Incorrect}} & \textbf{\textit{Correct}} & \textbf{\textit{Incorrect}} & \textbf{\textit{Correct}}\\
            \hline
        how many  &  $ 0.475$ & $ 0.468$  &  $ 0.372$ & $ 0.354$ &  $ 0.559 $ & $ 0.562 $\\
        what color  &  $ 0.454$ & $ 0.466$  &  $ 0.420$ & $ 0.517$ & $ 0.514 $ & $ 0.542 $\\
        what sport  &  $ 0.480$ & $ 0.584$ & $ 0.299$ & $ 0.342$ & $ 0.513 $ & $ 0.628 $\\
        what animal  &  $ 0.436$ & $ 0.544$ &  $ 0.257$ & $ 0.322$ & $ 0.488 $ & $ 0.623 $\\
        what time  &  $ 0.469$ & $ 0.405$   &  $ 0.333$ & $ 0.282$ & $ 0.528 $ & $ 0.492 $\\
        what brand  &  $ 0.440$ & $ 0.458$  &  $ 0.481$ & $ 0.511$ & $ 0.497 $ & $ 0.508 $\\
        what type/kind  &  $ 0.485$ & $ 0.537$  &  $ 0.382$ & $ 0.417$ & $ 0.544 $ &  $ 0.594 $ \\
        where  &  $ 0.501$ & $ 0.551$ & $ 0.380$ & $ 0.435$ &  $ 0.561 $ & $ 0.620 $\\
        which  &  $ 0.495$ & $ 0.529$ & $ 0.419$ & $ 0.414$ & $ 0.556 $ &$ 0.581 $\\
        what is/are the  &  $ 0.497$ & $ 0.543$  &  $ 0.436$ & $ 0.468$ & $ 0.559 $ & $ 0.605 $\\
        others  &  $ 0.480$ & $ 0.471$ &  $ 0.412$ & $ 0.370$& $ 0.549 $ & $ 0.550 $\\ \hline
    \end{tabular}}
    \caption{CLIPScore, UMIC, and PAC-S for correct and incorrect caption-like sentences generated for different question types of VQAv2.}
    \label{tab:clipscore_umic_vqa_question_types}
\end{table*}

\textbf{Score normalized results}: As shown in Table \ref{tab:vqa_tdiuc}, CLIPScore and UMIC assign higher average scores to incorrect captions compared to correct captions; however, PAC-S assigns higher average scores to correct captions. We conducted further analysis by examining the average scores assigned by these metrics for different answer types of the VQAv2 dataset (please refer to Table \ref{tab:vqa_answer_types} for detailed scores). Specifically, we observed that for the `yes/no' answer type, on average, all the metrics assign higher scores to incorrect captions. For the `number' answer type, only PAC-S was able to assign higher average scores to correct captions. However, for the `others' answer type, all the metrics assign higher average scores to correct captions.

For further investigation, we look at results for specific question types for VQAv2. As illustrated in Table \ref{tab:clipscore_umic_vqa_question_types}), for CLIPScore, we observe that incorrect captions received higher scores on average for three question types: `how many', `what time' and `others'. Also, UMIC assigns higher scores on average to incorrect captions for four question types: `how many', `what time', `which', and `others'. On the other hand, PAC-S assigns higher scores on average to incorrect captions for `what time' and `others' question types, suggesting \textbf{all metrics show poor performance for `what time' questions}, which is considered to be a hard question type. Moreover, \textbf{CLIPScore and UMIC show poor performance for `how many' questions.} Although PAC-S assigns higher average to correct captions over incorrect captions for `how many' question type, the gap between the absolute values of average scores for correct and incorrect captions for `how many' question is less than that for other question types.

\textbf{Controlled investigation to identify sensitivity to various factors:} Having established that these metrics struggle to distinguish the set of incorrect captions from the set of correct captions, in the following sections, we delve deeper into understanding the underlying reasons for their failure. To validate the comparisons made between different group means and ensure the reliability of our claims, we conducted a \textbf{t-test} for each comparison, using a p-value threshold of 0.01 (p-value < 0.01). Notably, all reported comparisons successfully satisfied this predetermined threshold, affirming the robustness of our statistical analyses.
    \subsection{Sensitivity to fine-grained errors}
        The primary objective of this section is to determine the sensitivity of these metrics to fine-grained errors. 
An incorrect caption is said to have ``fine-grained errors'' if it has high lexical overlap with a correct caption.
To obtain such pairs of correct and incorrect captions, we first generate incorrect captions corresponding to the questions for which ALBEF produced incorrect responses. Then, we generate correct captions using ground-truth answers for the same set of questions. We convert the questions and answers into captions using the method described in Section \ref{sec:datasets}. We excluded questions with yes/no answers from this study as we discuss them in Section \ref{sec:linguistic}. In total, we analyzed 38383 samples for this experiment. 

We quantify the \textbf{degree of lexical overlap} between a pair of correct and incorrect captions in our dataset by measuring the F1 score between them. The mean F1 score across all such pairs in our dataset is 0.725. To place this in context, we measure the F1 score between pairs of correct (human-written) and incorrect (generated by image captioning models) captions from the Composite dataset \cite{composite}, a widely-used dataset for evaluating image captioning metrics (see Appendix \ref{sec:F1_score_for_composite_dataset} for more details on F1 score computation for Composite dataset). The mean F1 score across all such pairs from the Composite dataset is 0.224, which is significantly lower than that for our dataset. This highlights the difficulty of our dataset making it suitable for stress testing the robustness of image captioning metrics.
\begin{table}
    \centering
    \resizebox{\linewidth}{!}
    {
    \begin{tabular}{lccc}
        \hline
        \textbf{Answer Type} & \textbf{CLIPScore} & \textbf{UMIC} & \textbf{PAC-S}\\
        \hline
        Ground Truth &  $ 0.479$ & $0.422$ & $0.542$\\
        Incorrect from ALBEF & $0.468$& $0.404$ & $0.535$\\ \hline
    \end{tabular}}
    \caption{CLIPScore, UMIC, and PAC-S comparison for caption-like sentences for incorrect answers generated by ALBEF model for VQAv2 and captions generated with its ground truth counterpart.}
    \label{tab:vqa_fine_grained_ALBEF}
\end{table}
As demonstrated in Table \ref{tab:vqa_fine_grained_ALBEF}, for all metrics, captions with ground truth answers received a higher average score compared to captions with fine-grained errors. Despite the higher average scores assigned to correct captions, the ranking results reveal that these metrics often fail to prioritize correct captions over incorrect ones. CLIPScore fails to rank correct captions above incorrect captions in 46.34\% of cases, while UMIC fails to do so in 45.99\% of cases. Also, PAC-S ranks incorrect captions over correct captions in 46.84\% of times. Thus, \textbf{all metrics show weak sensitivity to detecting fine-grained errors}.

We also report a \textbf{human baseline} 
for the task of distinguishing correct captions from the ones with fine-grained errors. We collected five human annotations for 2000 examples using the Amazon Mechanical Turk platform, each example consisting of an image, a correct caption and an incorrect caption. We asked humans to indicate the best matching description. Majority voting across the worker responses for each caption revealed humans fail to identify correct caption from incorrect caption in 15.4\% cases. This shows human performance is far better than the metrics’ performance which fail to rank correct captions above incorrect captions around 46\% of the time.


    \subsection{ Are metrics differently sensitive to different kinds of fine-grained errors?}
\begin{figure}[h]
  \centering
   \includegraphics[width=\linewidth]
   {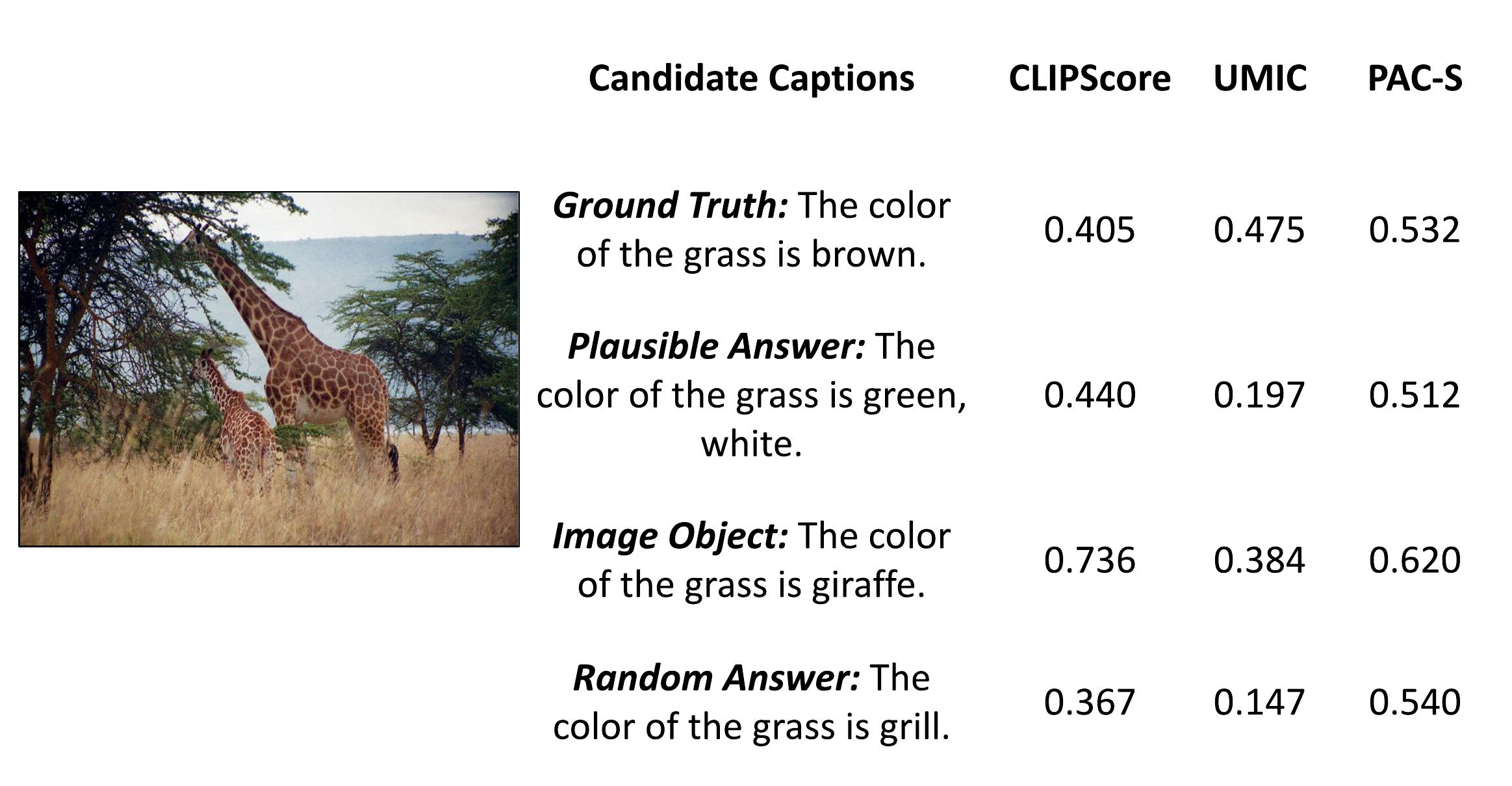}
   \caption{Captions from ground truth, plausible answer, an object from the image and a random asnwer of VQAv2.}
   \label{fig:diff_fine_grained_visualization}
\end{figure}
The main aim of this experiment is to assess if the metrics exhibit varying sensitivity to different types of fine-grained errors, in particular visual grounding errors and caption implausibility errors. To assess this, we generated three types of incorrect captions for each correct caption by replacing the ground-truth answer in the correct caption with: a plausible but incorrect answer (visual grounding error), an object found in the image (caption implausibility error), and a random answer (see Figure \ref{fig:diff_fine_grained_visualization} for an example and see Appendix \ref{sec:plausible_answers} for more details on plausible answers). 

For this experiment, we limited our investigation to the following question types: `what number is', `what time', `what color', and `what brand', as their answers are non-object entities and, therefore, are not present in the COCO Detection dataset. Thus, when constructing a sentence using an object in the image, we can be sure that it would result in an incorrect caption for the image. We analyzed 23841 sets of 4 captions each for this experiment.

\begin{table}
    \centering
    \resizebox{\linewidth}{!}
    {\begin{tabular}{lccc}
        \hline
        \textbf{Answer Type} & \textbf{CLIPScore} & \textbf{UMIC} & \textbf{PAC-S}\\
        \hline
        Ground Truth &  $0.501$ & $0.487$ & $0.576$\\
        Plausible & $ 0.474$ & $0.242$ & $0.527$\\
        Object from Image & $0.526$ & $0.354$ & $0.601$\\
        Random & $0.458$ & $0.275$ & $0.522$\\ \hline
    \end{tabular}}
    \caption{CLIPScore, UMIC, and PAC-S comparison for caption-like sentences from VQAv2 ground truth, plausible, object from image and random answers.}
    \label{tab:different_fine_grained}
\end{table}


As illustrated in Table \ref{tab:different_fine_grained}, the score difference between the correct captions and the captions with implausibility errors is significantly smaller than the difference between the correct captions and the captions with visual grounding errors. This indicates that the metrics exhibit \textbf{lower sensitivity} to caption \textbf{implausibility errors} and \textbf{higher sensitivity} to \textbf{visual grounding errors}. Notably, both CLIPScore and PAC-S assigned higher average scores to captions with implausibility errors compared to ground truth answers, and only UMIC assigned higher average score to captions with ground truth answers. In the following sections, we further examine the sensitivity of the metrics to various visual and linguistic aspects.
    \subsection{Visual Aspects}
        \label{sec:visual}
       In this section, our objective is to assess the sensitivity of the metrics to the size and number of objects mentioned in the caption. Importantly, we would like to highlight that our focus is on analyzing how the size and number of objects mentioned in captions affect metric robustness and sensitivity. We refrain from making value judgments about whether these effects are good or bad.
        \subsubsection{\textbf{Sensitivity to the number of object mentions in the caption}}
\begin{figure}[h]
  \centering
   \includegraphics[width=\linewidth]
   {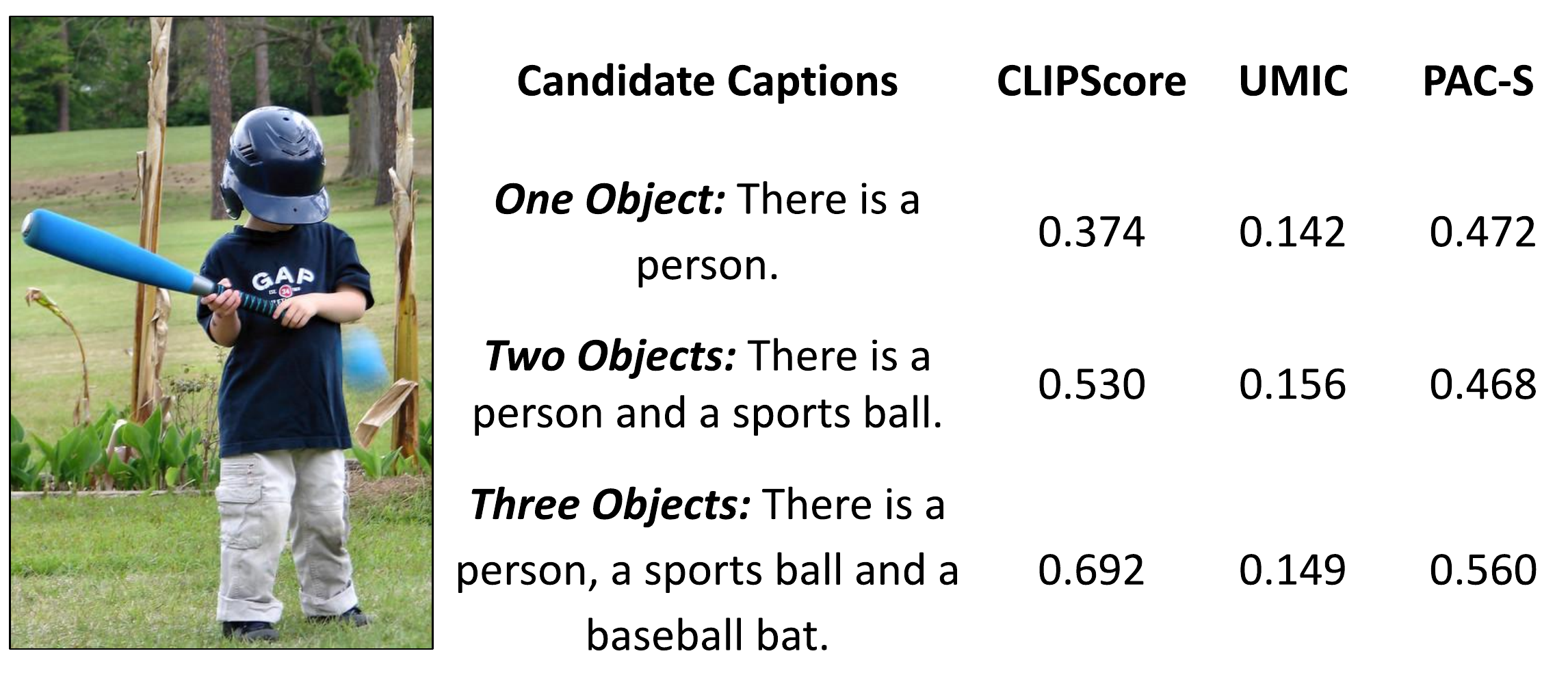}
   \caption{Captions referring to different number of objects from the image.}
   \label{fig:num_objects}
\end{figure}

In this section, we aim to evaluate the sensitivity of the metrics to the number of objects mentioned in the caption. To conduct this evaluation, we filter images from COCO Detection dataset \cite{coco} having a minimum of three object tags and randomly select three object tags for each image and utilize their corresponding object names to form sentences, depicting one, two, and three objects presented in the image (see Figure \ref{fig:num_objects}). We analyzed 19412 images for this experiment.
\begin{table}
    \centering
    \resizebox{\linewidth}{!}
        {\begin{tabular}{lccc}
            \hline
            \textbf{Number of Objects} & \textbf{CLIPScore} & \textbf{UMIC}  & \textbf{PAC-S}\\
            \hline
            One Object &  $0.449$ & $0.205$ & $0.500$\\
            Two Objects & $0.512$ & $0.212$ & $0.540$\\
            Three Objects & $0.561$ & $0.195$ & $0.578$\\
            Shuffled One Object & $0.445$ & $ 0.139$ & $0.503$\\
            Shuffled Two Objects & $0.499$ & $0.148$ & $0.541$\\ 
            Shuffled Three Objects & $0.540$  & $0.169$ & $0.576$\\ \hline
        \end{tabular}}
    \caption{CLIPScore, UMIC, and PAC-S comparison for sentences with various number of objects name, and their shuffled counterparts.}
    \label{tab:num_tags}
\end{table}
As presented in the first three rows of Table \ref{tab:num_tags}, CLIPScore and PAC-S scores for captions with three objects are significantly higher than for captions with two objects. Also, captions with two objects score significantly higher than those with one object. In contrast, for UMIC, captions with one, two, and three objects received average scores of 0.205, 0.212, and 0.195, respectively. Although the t-test indicated statistically significant differences between scores across different object counts, the gap between absolute score values is smaller for UMIC than for CLIPScore and PAC-S. In conclusion, \textbf{CLIPScore and PAC-S display a heightened sensitivity to the number of image-relevant objects mentioned in the caption, while UMIC shows limited sensitivity towards this factor}.
        \subsubsection{\textbf{Sensitivity to size of objects mentioned in the caption}}
\begin{figure}[b]
  \centering
   \includegraphics[width=\linewidth]
   {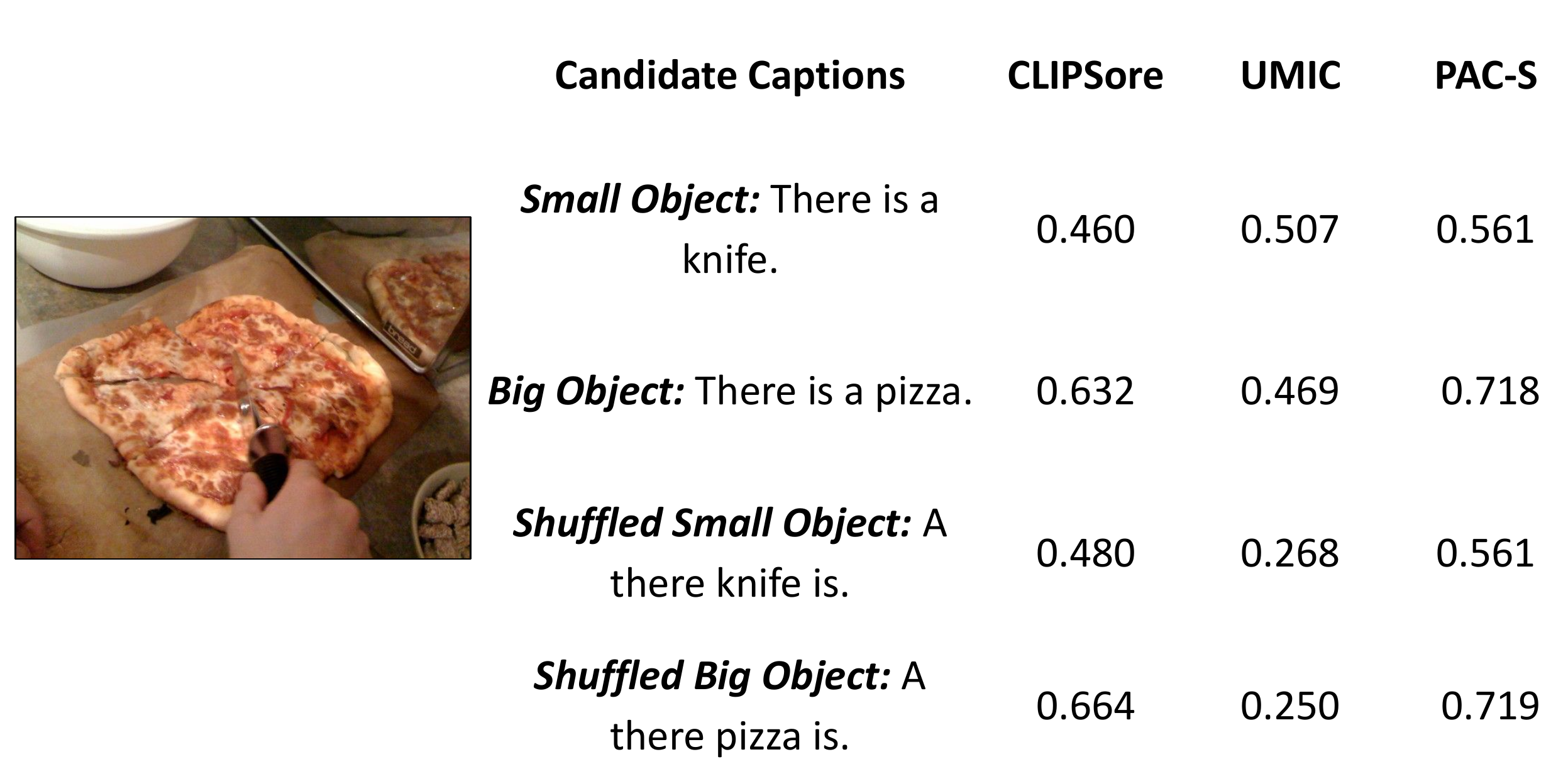}
   \caption{Captions referring to small and large area of the image and their shuffled counterparts.}
   \label{fig:object_size}
\end{figure}
In this experiment, our primary goal is to examine the effect of object size mentioned in captions on the metrics. To achieve this, we utilize the COCO Detection dataset \cite{coco} to select one small and one large object from the same image with a noticeable difference in the area (see Figure \ref{fig:object_size} for an example and for detailed explanation see Appendix \ref{sec:detailed_pick_object_size}.). As a result, we selected 24610 images for further analysis.
\begin{table}
    \centering
    \resizebox{\linewidth}{!}
        {\begin{tabular}{lccc}
            \hline
            \textbf{Object Size} & \textbf{CLIPScore} & \textbf{UMIC }& \textbf{PAC-S}\\
            \hline
            Small Object &  $0.396$ & $0.317$ & $0.492$\\
            Big Object & $0.434$ & $0.232$ & $0.580$ \\
            Shuffled Small Object & $0.390$ & $0.205$ & $0.495$\\
            Shuffled Big Object & $0.436$ & $0.170$ & $0.590$\\ \hline
        \end{tabular}}
    \caption{CLIPScore, UMIC, and PAC-S comparison for captions referring to small and a big objects in the image, and their shuffled counterparts.}
    \label{tab:tag_size}
\end{table}

As demonstrated in the first two rows of Table, \ref{tab:tag_size}, for CLIPScore and PAC-S, captions with smaller objects received a lower average score than those with bigger objects. On the other hand, UMIC assigned a higher average score to captions with smaller objects compared to captions with bigger objects. Overall, \textbf{all metrics demonstrate sensitivity to the size of image-relevant objects mentioned in the caption}.
    \subsection{Linguistic Aspects}
        \label{sec:linguistic}
        \subsubsection{\textbf{Sensitivity to negation}}
To assess the ability of metrics to distinguish between correct captions and their negated versions, we created 80530 captions-like sentences by using the questions with `yes' or `no' ground-truth answers from the validation split of VQAv2. Additionally, we generated negated captions by negating the ground truth answer.

For CLIPScore, correct captions received a higher score of $0.457$, and their negated versions got $0.450$ on average. For UMIC, correct captions received a higher average of $0.359$, and their negated versions got $0.335$ on average. Correct captions received a higher average of $0.556$ for PAC-S, and their negated versions got $0.548$ on average. Although the correct captions scored statistically significantly higher than the negated ones, CLIPScore, UMIC, and PAC-S ranked the negated caption above the correct caption incorrectly in 41.36\%, 44.24\%, and 41.83\% of cases, respectively. Thus, \textbf{all metrics exhibit a weak understanding of negation}.
        \subsubsection{\textbf{Sensitivity to the sentence structure}}
            To evaluate the sensitivity of the metrics to sentence structure, we generated 214354 caption-like sentences with VQAv2 ground truth answers and then shuffled them. For CLIPScore, correct captions received $0.469$, and their shuffled version got $0.450$ on average. For UMIC, correct captions received $0.400$, and their shuffled version got $0.211$ on average. Correct captions received $0.548$ for PAC-S, and their shuffled version got $0.539$ on average. Despite higher average scores assigned to correct captions, the ranking results reveal that CLIPScore fails to rank the correct caption higher than the shuffled one in 34.32\% of cases, contrasting with UMIC, where this occurs in only 9.18\% of cases. Additionally, PAC-S falls short, assigning a higher score to the correct caption than the shuffled one in 43.05\% of cases. This indicates that \textbf{UMIC is more responsive to the structure of the sentence compared to CLIPScore and PAC-S}.
    \subsection{Visio-Linguistic Aspects}
        \subsubsection{\textbf{Sentence Structure versus Visual Aspects}}
        \label{sec:structure_vs_visual}
            In order to compare the sensitivity of metrics to sentence structure and object size, we conducted a sentence shuffling experiment using captions that contained objects of varying sizes, as described in Section \ref{sec:visual}. We shuffle both big and small object captions in the same order (see Figure \ref{fig:object_size}). As shown in Table \ref{tab:tag_size}, our results demonstrate that CLIPScore and PAC-S assign the highest scores to captions referring to a larger area of the image, regardless of whether they are shuffled or not. In contrast, UMIC exhibits the opposite trend, with the highest scores assigned to correct (i.e., unshuffled) sentences, regardless of the size of the objects mentioned in the captions. This highlights that \textbf{UMIC is more sensitive to sentence structure than the size of the objects mentioned in the caption, whereas for CLIPScore and PAC-S, the behavior is just the opposite}.

To compare the sensitivity of metrics to sentence structure and the number of object mentions, we conducted a sentence shuffling experiment using captions that varied in the number of object mentions. As shown in Table \ref{tab:num_tags}, UMIC assigns the lowest scores to shuffled captions, regardless of the number of objects mentioned in the captions. \textbf{This indicates that UMIC prioritizes sentence structure over the number of object mentions}. In contrast, CLIPScore and PAC-S assign the highest scores to captions with three objects, regardless of whether they are shuffled or not. Similarly, the captions with two objects have the second highest CLIPScore and PAC-S, regardless of the correctness of the sentence structure. This reveals that \textbf{CLIPScore and PAC-S places greater importance on the number of object mentions than the sentence structure}.

\section{Conclusion and Discussion}
In conclusion, recently proposed reference-free image captioning evaluation metrics are far from perfect; they cannot distinguish an incorrect caption from a correct caption when the difference between them is fine-grained. The sensitivity of CLIPScore, UMIC, and PAC-S varies across different error types: they are less affected by plausibility errors yet more by visual grounding errors. All metrics struggle with understanding negation. All metrics are influenced by the size of the relevant objects mentioned in the caption, and CLIPScore and PAC-S also responds to the number of object mentions. UMIC is responsive to sentence structure, while CLIPScore and PAC-S disregards it often. Moreover, UMIC prioritizes sentence structure over the number and size of objects mentioned in the caption; in contrast CLIPScore and PAC-S prioritize the object size and number of object mentions over sentence structure.

Our primary contribution is highlighting specific areas where reference-free metrics exhibit limitations. The root cause of these limitations is traced to the insufficient fine-grained understanding of the CLIP and UNITER models upon which these reference-free metrics rely. In order to improve the reference-free metrics, we believe that underlying models need to become better at fine-grained understanding of objects, attributes, relationships etc., so that they can better distinguish fine-grained differences between captions. Promising avenues for enhancing this understanding include exploring object-centric representations \cite{Object-Centric, Multi-object, Monet} and incorporating training with hard negatives \cite{yuksekgonul2023when, zhang2023contrasting, bugliarello2023weaklysupervised}, allowing the model to learn to discern fine-grained differences and errors. Given the restricted fine-grained understanding of the underlying models shaping these metrics, caution is advised when employing them as evaluation metrics for image captioning.

\section*{Limitations}
    As a limitation, it is important to consider that responses marked as incorrect may not always be incorrect due to the stringent nature of VQA evaluation metrics \cite{vqa_strict_eval}. Our approach does not account for this factor. However, for our experiments, since we fine-tune ALBEF for each domain, the risk of this issue is low. To get a quantitative sense, we randomly sampled 100 incorrect answers (as deemed by the VQA automatic metric) generated by ALBEF for VQAv2, and in only 10\% of cases, the answer was actually correct (as deemed by an expert human). Furthermore, it is important to note that we do not account for the saliency of objects mentioned in the caption, which could be a confounding factor in our evaluation.

\section*{Ethics Statement}
    To enhance transparency and explainability, we conducted experiments aimed at shedding light on the evaluation process of the metric. By doing so, we aimed to provide insights and explanations that enable users to better comprehend and trust the metric's evaluations. Furthermore, we evaluated the robustness of the metrics, contributing towards the development of less biased evaluation metrics.

While we assess various aspects of existing metrics, it is important to note that our evaluation does not specifically examine metrics’ potential biases across different demographics, including gender or race.
While our research does not include an explicit experiment on bias perpetuation or amplification, we strongly encourage future studies to investigate how metrics may interact with biases present in datasets. This research direction is crucial in developing metrics that are less biased and more inclusive towards diverse demographics.

\section*{Acknowledgements}
We express our gratitude to Stefan Lee for providing constructive feedback. The technical support extended by the Mila IDT team in managing the computational infrastructure is greatly appreciated. The authors acknowledge the material support of NVIDIA in the form of computational resources. Throughout this project, Aishwarya Agrawal received support from the Canada CIFAR AI Chair award.

\bibliography{anthology,custom}
\bibliographystyle{acl_natbib}

\appendix

\section{Appendix}
\label{sec:appendix}

\subsection{Generating Caption-like Sentences}
\label{sec:generate_caption_like_senteces}

To generate caption-like sentences from each question and answer pair of VQA datasets, we utilize pretrained GPT-J \cite{wang2021gptj} in a few-shot manner. To accomplish this, we first constructed a support examples dataset using the VQAv2 \cite{goyal2017making} training split. For each of the sixty-four predefined question types in the VQAv2 dataset, we randomly selected four examples from the VQAv2 training split. Then, we transformed both the questions and answers into single sentences, which we wrote ourselves. When generating captions for VQAv2 validation split, we first match the question type to one of the predefined sixty-four question types. Then, we select four support examples associated with that question type and prompt GPT-J to generate a transformed sentence. If the question type does not match any of our predefined question types, we randomly select eight support examples from the entire pool of support examples. Please see Figure \ref{fig:caption_like} and note that we visualized a 2-shot prompt for simplification.

\subsection{Comparison with FOIL, Winoground and ARO}
\label{sec:appendix_other_dataset}

\begin{itemize}
    \item FOIL: The distinction between our dataset and the FOIL dataset lies in their respective approaches to altering captions. While FOIL primarily focuses on changing nouns in MS-COCO captions, encompassing 73 out of the 91 MS-COCO categories, our setup, utilizing the VQA dataset, allows for a more diverse analysis. In our study, we go beyond changing nouns and explore variations in captions related to colors, time, count, and more. Notably, even in terms of nouns, our dataset exhibits greater diversity as we are not constrained to object types present in MS-COCO annotated categories.
    \item ARO: ARO dataset incorporates tests focusing on attribution, relations, and order. In the attribution test, distinctions are drawn between phrases like "The paved road and the white house." and "The white road and the paved house.". Meanwhile, the relation test explores understanding relationships, as seen in examples like "The horse is eating the grass." and the contrasting, implausible statement "The grass is eating the horse.". As shown by \cite{hsieh2023sugarcrepe}, the hard-negative captions present in these benchmarks are easily identifiable by vision-language models as they are out-of-distribution (OOD) w.r.t the training data seen by the language encoder in these models. While our correct and incorrect pairs of captions are both plausible sentences where only the incorrect caption exhibits a fine-grained error that stems from a lack of precise visual grounding.
    \item Winoground: Winoground dataset is meticulously curated by humans specifically for testing visio-linguistic compositionality. While it maintains a high level of quality, it comprises only 1600 samples, which, regrettably, is insufficient for robust statistical analyses. Furthermore, it lacks detailed annotations for aspects such as color, time, and counting in comparison to VQAv2. Importantly, as indicated by \cite{diwan-etal-2022-winoground}, this dataset introduces challenges that go beyond fine-grained understanding, including issues like out-of-domain challenges and ambiguous captions. These challenges significantly confound the study's results.
\end{itemize}

\subsection{F1 score computation for the Composite Dataset}
\label{sec:F1_score_for_composite_dataset}
We calculated the F1 score between the human-written correct captions and model-generated incorrect captions in the Composite dataset \cite{composite}. We used the captions generated by the Karparthy model \cite{karpathy} as they were better in quality. In the Composite dataset, each model-generated caption has an associated correctness score (provided by humans) ranging from 1 (‘The description has no relevance to the image’) to 5 (‘The description relates perfectly to the image’).
For our F1 score computation, we considered all captions with score less than or equal to 4 as incorrect captions.

\subsection{Plausible Answers}
\label{sec:plausible_answers}
To generate plausible captions for each question type, we first compiled a list of plausible answers derived from the ground truth multiple-choice answer of the same question type in the validation split of VQAv2. Subsequently, an answer was randomly selected from this list of plausible answers. This chosen answer was used to replace the ground truth answer in the original caption, thus generating a plausible alternate caption.

\subsection{Picking a large and small object from the image}
\label{sec:detailed_pick_object_size}
In this experiment, our primary objective is to investigate how the object size mentioned in captions affects the scores assigned by CLIPScore and UMIC. To select small and large objects that are distinctly different in size, we could sort the objects by their associated area in the COCO Detection dataset. However, this approach may not always yield accurate results because multiple objects with the same name may appear in an image. For instance, if there are two cars in an image, one smaller but further away and the other larger but closer, sorting by area would lead to incorrect identification of the smallest and largest objects. This would result in identical captions for both objects, such as ``There is a car." which is not ideal for comparison. 

To overcome this issue, we added up the area of all object categories with the same name and sorted the total areas of each object category in the image. We then calculated the difference between the areas associated with the largest and smallest categories. If the difference exceeded our threshold, we selected those objects for analysis. As a result, we selected 24610 images for further analysis (See Figure \ref{fig:object_size}).

\subsection{Computational Resources}
In all experiments detailed in this paper, we employed a single NVIDIA Quadro RTX 8000 with 48 GB GDDR6 GPU Memory. Specifically, for the primary task of generating caption-like sentences from the VQAv2 dataset, we performed inference using the GPT-J model with 6 billion parameters, executing the process over a duration of 24 hours.

\subsection{Dataset Terms of Use}
We will distribute our datasets (both generated with caption template and QA to caption conversion method) under the Creative Commons Attribution 4.0 License. It is noteworthy to mention that this licensing choice aligns with the terms of use governing both the COCO and VQAv2 datasets, foundational to the creation of our datasets.

\subsection{Editorial Assistance}
We would like to disclose that ChatGPT was utilized for refining the language and structure of this academic paper. While the primary content and research remain the work of the authors, the assistance provided by ChatGPT was limited to the improvement of writing quality.

\end{document}